\documentclass{article}

\usepackage[preprint]{corl_2023} 

\usepackage[numbers]{natbib}
\usepackage{multicol}
\usepackage{commath}
\usepackage{graphicx}
\usepackage{svg}
\usepackage{todonotes}
\usepackage{subcaption}
\usepackage{float}

\usepackage{makecell, caption, booktabs, multirow, siunitx}

\usepackage{lipsum, fancyvrb}

\usepackage{stfloats}

\newcommand{\OurMethod}{FOCUS}

\def\TheTitle{\OurMethod{}: Object-Centric World Models \\ for Robotics Manipulation}

\title{\TheTitle{}}

\author{
  Stefano Ferraro$^{*}$ \qquad
  Pietro Mazzaglia$^{*}$ \qquad
  Tim Verbelen \qquad
  Bart Dhoedt 
  \vspace{0.25cm} \\
  Ghent University\\ 
  Belgium\\
  \texttt{stefano.ferraro@ugent.be} 
}

\begin{document}
\maketitle

\def\thefootnote{*}\footnotetext{These authors contributed equally to this work}\def\thefootnote{\arabic{footnote}}


\begin{abstract}
Understanding the world in terms of objects and the possible interplays with them is an important cognition ability, especially in robotics manipulation, where many tasks require robot-object interactions. However, learning such a structured world model, which specifically captures entities and relationships, remains a challenging and underexplored problem. To address this, we propose \OurMethod{}, a model-based agent that learns an object-centric world model. Thanks to a novel exploration bonus that stems from the object-centric representation, \OurMethod{} can be deployed on robotics manipulation tasks to explore object interactions more easily. Evaluating our approach on manipulation tasks across different settings, we show that object-centric world models allow the agent to solve tasks more efficiently and enable consistent exploration of robot-object interactions. Using a Franka Emika robot arm, we also showcase how \OurMethod{} could be adopted in real-world settings. 

\textbf{Project website:} \enskip \url{https://focus-manipulation.github.io/}
\end{abstract}

\section{Introduction}


For robot manipulators, the tasks we perform as humans are extremely challenging due to the high level of complexity in the interaction between the agent and the environment. In recent years, deep reinforcement learning (RL) has shown to be a promising approach for dealing with these challenging scenarios \cite{Levine2016DeepVMPolicies, OpenAI2019Rubik, Kalashnikov2018QTOpt, Lu2021AWOpt, Lee2021Stacking}.

Among RL algorithms, model-based approaches, which allow learning behaviors using trajectories imagined with the model, {a.k.a. "learning} in imagination" \cite{Hafner2020Dreamer}, aspire to provide greater data efficiency, compared to the model-free counterparts \cite{Fujimoto2018TD3, Haarnoja2018SAC}. Adopting world models \cite{Ha2018WM, Hafner2021DreamerV2}, {i.e. generative} models that learn the environment dynamics by reconstructing the agent's observations, model-based agents have shown impressive performance across several domains \cite{Hafner2021DreamerV2, Rajeswar2023MasteringURLB, Hafner2023DreamerV3}, including real-world applications, such as robotics manipulation and locomotion \cite{Wu2022DayDreamer, Seo2023MV-MWM}.

However, world models that indistinctly reconstruct all information in the environment can suffer from several failure modes. For instance, in visual tasks, they can ignore small, but important features for predicting the future, such as little objects \cite{Seo2022MaskWM}, or they can waste most of the model capacity on rich, but potentially irrelevant features, such as static backgrounds \cite{Deng2021DreamerPro}. In the case of robot manipulation, this is problematic because the agent strongly needs to acquire information about the objects to manipulate in order to solve tasks. 

Another challenge in RL for manipulation is engineering reward functions, able to drive the agent's learning toward task completion, as attempting to design dense reward feedback easily leads to faulty reward designs \cite{Amodei2016RewardHacking, Clark2016FaultyReward, Krakovna2020SpecificationGaming, Popov2017LegoStacking}. One solution is to adopt sparse reward feedback, providing a positive reward only for successful task completion. However, these functions are challenging to optimize with RL, due to the difficulty of finding rewards in the environment and thus require appropriate exploration strategies, for which previous work has resorted to artificial curiosity mechanisms, encouraging the agent to search for novel interactions with the environment \citep{Oudeyer2007IntrinsicMotivationSystems, Schmidhuber1991CuriousModel}. 

Humans tend to develop a structured mental model of the world by interacting with objects, registering specific features associated with objects, such as shape, color, etc \cite{Hawkins2017CorticalColumns}. Since infancy, toddlers learn this by actively engaging with objects and manipulating them with their hands, discovering object-centric views that allow them to build an accurate mental model \cite{smith2018toddlers,slone2019toddlers}. 




Inspired by the principle that objects should be of primary importance in the agent's world model, we present \textbf{\OurMethod{}}, a model-based RL agent that learns an object-centric representation of the world and is able to explore object interactions. 

\textbf{Contributions} Our contributions can be summarized as:
\begin{itemize}
    \setlength\itemsep{0pt}
    \item An object-centric latent dynamics model, which learns the dynamics of the environment, while discriminating object representations into distinct latent vectors. (Section \ref{sec:obj_model});
    \item An object-centric exploration strategy, which encourages interactions with the objects, by maximizing the entropy of the latent object representation. (Section \ref{sec:obj_expl});
    \item We show that object-centric world models allow solving dense reward manipulation tasks faster (Section \ref{sec:dense_tasks});
    \item We demonstrate that \OurMethod{}' exploration strategy strongly fosters interaction with objects and facilitates adaptation to sparse reward tasks, after an exploration stage (Section \ref{sec:expl_tasks});
    \item We analyze the effectiveness of the object-centric model, showing that it captures crucial information for manipulation, both in simulation and in real-world scenarios (Section \ref{sec:decoder}).
\end{itemize}

\section{Background}

\textbf{Reinforcement learning.} In RL, the agent receives inputs $x$ from the environment and can interact through actions $a$. The objective of the agent is to maximize the discounted sum of rewards $\sum_t \gamma^tr_t$, where $t$ indicates discrete timesteps.
To do so, RL agents learn an optimal policy $\pi(a|x)$ outputting actions that maximize the expected cumulative discounted reward over time, generally estimated using a critic function, which can be either a state-value function $v(x)$ or an action-value function $q(x,a)$ \cite{Haarnoja2018SAC, Fujimoto2018TD3}.
World models \cite{Ha2018WM} additionally learn a generative model of the environment, capturing the environment dynamics into a latent space, which can be used to learn the actor and critic functions using imaginary rollouts \cite{Hafner2021DreamerV2, Hafner2023DreamerV3} or to actively plan at each action \cite{Schrittwieser2020MuZero, Hansen2022TDMPC, Rajeswar2023MasteringURLB}, which can lead to higher data efficiency in solving the task.  




\textbf{Exploration} Solving sparse-reward tasks is a hard problem in RL. Inspired by artificial curiosity theories~\cite{Schmidhuber1991CuriousModel, Oudeyer2007IntrinsicMotivationSystems}, several works have designed exploration strategies for RL~\cite{Pathak2017ICM, Mazzaglia2022LBS, Rajeswar2021TouchCuriosity}. Other explorations strategies that have shown great success over time are based upon maximizing uncertainty~\cite{Pathak2019Disagreement, Sekar2020P2E},
or the entropy of the agent's state representation~\cite{Liu2021APT, Seo2021RE3, Mutti2021MEPOL}. One issue with exploration in visual environments is that these approaches can be particularly attracted by easy-to-reach states that strongly change the visual appearance of the environment \cite{Burda2018LS}. For robotics manipulation this can cause undesirable behaviors, e.g. a robot arm exploring different poses in the proximity of the camera but ignoring interactions with the objects in the workspace \cite{Rajeswar2023MasteringURLB}.

\textbf{Object-centric representations.} Decomposing scenes into objects can enable efficient reasoning over high-level building blocks and ensure the agent hones in on the most relevant concepts \cite{Dittadi2022Generalization}. Several 2D object-centric representations, based on the principle of representing objects separately in the model, have been studied in the last years \cite{Locatello2020SlotAttention, Greff2020IODINE, Burgess2019MONet, Nakano2023InteractionOC}. Inspired by the idea that such representations could help exploit the underlying structure of our control problem \cite{Diuk2008OCRL, Dittadi2022Generalization}, we propose a world model with an object-centric structured representation \cite{Kipf2020CSWM} that we show could strongly aid robotics manipulation settings. 

\section{Method}


\subsection{Object-centric World Model}
\label{sec:obj_model}
The agent observes the environment through the inputs $x_t=\{o_t, q_t\}$ it receives at each interaction, where we can distinguish the (visual) observations $o_t$, e.g. camera RGB and depth, from the proprioceptive information $q_t$, i.e. the robot joint states and velocities. This information is processed by the agent through an encoder model, $e_t=f_\phi(x_t)$, which can be instantiated as the concatenation of the outputs of a CNN for high-dimensional observations and an MLP for low-dimensional proprioceptive inputs.

The world model aims to capture the temporal dynamics of the inputs into a latent state $s_t$. In previous work, this is achieved by reconstructing all the inputs using an observation decoder. With \OurMethod{}, we are interested in providing a structured world model representation that separates the object-specific information into separate latent representations $s_t^{obj}$. For this reason, we instantiate two object-conditioned components: an \textit{object latent extractor} and an \textit{object decoder}.

\begin{figure*}[t]
  \centering
  \includegraphics[width=0.9\textwidth]{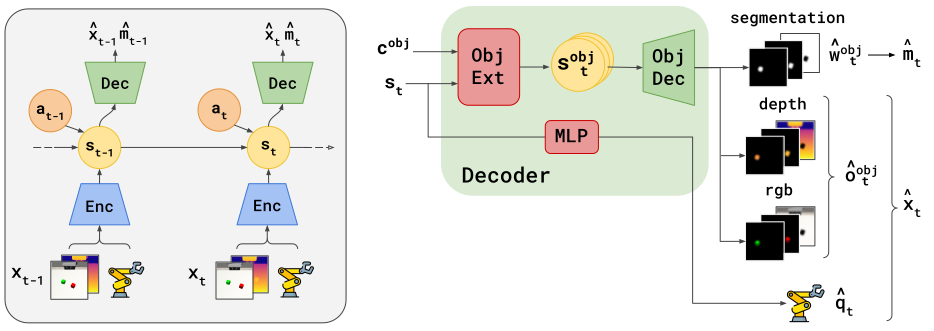}
  \caption{\textbf{\OurMethod{}.} Visual description of the object-centric world model learned by \OurMethod{}. \textit{(Left)} Overall composition of the model, showing the information flow through the different components. \textit{(Right)} Detailed view of the decoder, including the object latent extractor and object decoder. }
  \label{fig:architecture}
\end{figure*}

\textbf{Models.} Overall, the learned world model, which we illustrate in Fig.~\ref{fig:architecture} is composed of the following components:
\begin{align}
\begin{split}
\textrm{Encoder:} & \quad e_t = f_\phi(x_t), \\ 
\textrm{Posterior:} & \quad p_{\phi}(s_{t+1}|s_t, a_t, e_{t+1}), \\ 
\textrm{Prior:} & \quad p_{\phi}(s_{t+1}|s_t, a_t), \\
\end{split}
\begin{split}
\textrm{Proprio decoder:} & \quad p_{\theta}(\hat{q_t}|s_t), \\
\textrm{Object latent extractor:} & \quad p_{\theta}(s_t^{obj}|s_t, c^{obj}), \\
\textrm{Object decoder:} & \quad p_{\theta}(\hat{o}_t^{obj}, w_t^{obj}|s_t^{obj}),
\end{split}
\end{align}
where $\phi$ indicates the set of learnable parameters.

We adopt a recurrent state-space model (RSSM) \cite{hafner2019planet} for the dynamics components, i.e. prior and posterior, which extracts a latent state $s_t$ made of a deterministic and a stochastic component.
Proprioceptive information $\hat{q_t}$ is decoded out of the latent state $s_t$, using an MLP. For each object in the scene, the \textit{object latent extractor} receives the world model's latent state $s_t$ and a vector identifying the object $c^{obj}$, and extracts an object-centric latent $s_t^{obj}$. 
Given such object latent vectors, the \textit{object decoder} outputs two values: the object-related observation information $\hat{o}_t^{obj}$, and the segmentation ``unnormalized weights'' $w^{obj}_t$. 

The object-related information $\hat{o}_t^{obj}$ is a masked version of the visual inputs $o_t$ where all the irrelevant information to the object is masked out, using a segmentation mask. The one-dimensional ``unnormalized weights'' $w^{obj}_t$ represent object-specific per-pixel logits for segmenting the objects. The overall scene segmentation is obtained by applying a softmax among all object weights, where each pixel is assigned to the object that outputs the highest weight.

In practice, the object latent extractor is instantiated as an MLP while the object decoder is instantiated as a transposed CNN. For the object-conditioning vector $c^{obj}$ we adopt a one-hot vector identifying the object instance.


\textbf{Objective.} The model is trained end-to-end by minimizing the following loss:
\begin{equation}
    \mathcal{L_\textrm{wm}} = \mathcal{L}_\textrm{dyn} + \mathcal{L}_\textrm{proprio} + \mathcal{L}_\textrm{obj}.
\end{equation}
The dynamics minimizes the Kullback–Leibler (KL) divergence between posterior and prior:
\begin{equation}
    \mathcal{L_\textrm{dyn}} = D_\textrm{KL}[p_{\phi}(s_{t+1}|s_t, a_t, e_{t+1}) || p_{\phi}(s_{t+1}|s_t, a_t)].
\end{equation}
The proprioceptive decoder learns to reconstruct proprio states, by minimizing a negative log-likelihood (NLL) loss:
\begin{equation}
\begin{split}
    \mathcal{L}_\textrm{proprio} = -\log p_\theta(\hat{q_t}|s_t) 
\end{split}
\end{equation}
The object decoder learns to reconstruct object-centric information, outputting ``object weights'' for the segmentation mask and reconstructing object-related observations. The observation is masked via an object-specific mask $m^i_t$, to focus only on the $i$-th object information in the loss. The object decoder loss can be expressed as:
\begin{equation}
\begin{split}
    \mathcal{L_\textrm{obj}} = -\log \underbrace{p(\hat{m_t})}_\textrm{mask}  -\log \sum^{N}_{i=0} \underbrace{m_t^i p_{\theta}(\hat{x_t^i}|s_t^i)}_\textrm{masked reconstruction}
\end{split}
\end{equation}
where the overall segmentation mask is obtained as:  
\begin{equation}
\hat{m_t}=\textrm{softmax}(w^1_t, ..., w^N_t)   
\end{equation}
with $N$ being the number of all object instances present in the scene\footnote{The scene, with objects masked out, is also considered a "special object".}. By minimizing the \textit{masked reconstruction} NLL, the object-decoder ensures that each object latent $s^i$ focuses on capturing only its relevant information, as the reconstructions obtained from the latent are masked per object. Furthermore, objects compete for occupying their correct space in the scene (in pixel space), through the \textit{mask} loss. Thus, each object's latent vector is aware of the object's positioning in the scene.

In addition to the above components, when the agent intends to solve reward-supervised tasks, a reward predictor $p_\phi(\hat{r}_t|s_t)$ is learned to minimize the loss $\mathcal{L_\textrm{rew}}=-\log p_\phi(\hat{r}_t|s_t)$. Further details about the model architecture, the objective, and the optimization process are provided in Appendix.


\subsection{Object-centric Exploration}
\label{sec:obj_expl}
State maximum entropy approaches for RL \cite{Mutti2021MEPOL, Seo2021RE3, Liu2021APT} learn an environment representation, on top of which they compute an entropy estimate that is maximized by the agent's actor to foster exploration of diverse environment states. 
Given our object-centric representation, we can incentivize exploration towards object interactions and discovery of novel object views, by having the agent maximize the entropy over the object latent vectors. 

In order to estimate the entropy value over training batches, we apply a K-NN particle-based estimator \cite{Singh2003NNPBE} on top of the object latent representation. By maximizing the overall entropy, with respect to all objects in the scene, we derive the following reward for object-centric exploration: 
\begin{equation}
    r_\textrm{expl} = \sum^{N}_{obj=0} r_\textrm{expl}^{obj}, \quad
    \textrm{where} \quad r_\textrm{expl}^{obj}(s) \propto \sum^{K}_{i=1} \log \norm{s^{obj} - s^{obj}_i}_2
\label{eq:expl_rew}
\end{equation}
where $s^{obj}$ is extracted from $s$ using the object latent extractor, conditioned by the vector $c^{obj}$, and $s^{obj}_i$ is the $i$-th nearest neighbor to $s^{obj}$. 

Crucially, as we use the world model to optimize actor and critic networks in imagination \cite{Hafner2021DreamerV2}, the latent states in Equation \ref{eq:expl_rew} are states of imaginary trajectories, generated by the world model by following the actor's proposed actions.


\begin{figure}[b]
    \centering
    \includegraphics[width=\textwidth]{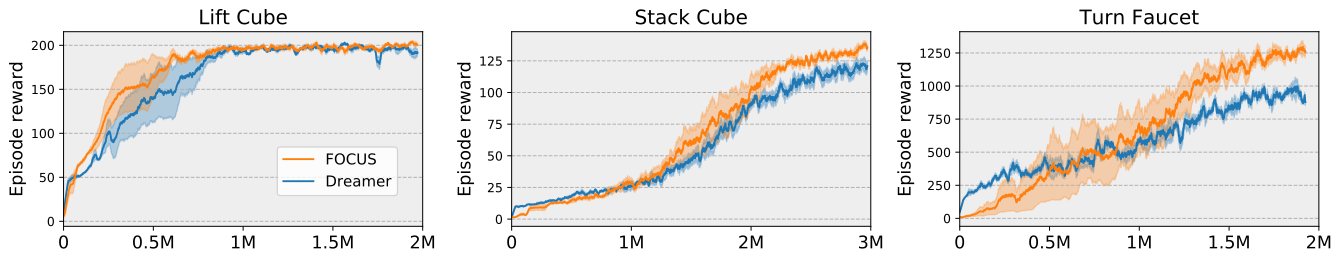}
    \caption{\textbf{Manipulation tasks.} Comparing \OurMethod{} and Dreamer over dense supervised manipulation tasks, to assess the quality of the world model. (3 seeds)}
    \label{fig:task_results}
\end{figure}

\section{Experiments}


We empirically evaluate \OurMethod{} aiming to answer the following questions: 
\begin{itemize}
    \item \textbf{Does \OurMethod{} object-centric world model enable faster learning of robotics manipulation tasks?} (Section \ref{sec:dense_tasks}) We test FOCUS against the DreamerV2 world model-based agent on a set of supervised dense reward tasks.
    \item \textbf{Can our object-driven exploration lead to more meaningful exploration, i.e. exploring interactions with the objects in the scene?} (Section \ref{sec:expl_tasks}) We deploy different exploration strategies across several environments and compare them using exploration metrics. Then, we evaluate whether the exploration stage was useful, by fine-tuning the exploration approaches on sparse-reward tasks in the same environments.
    \item \textbf{Does the learned world model captures crucial information for manipulation better?} (Section \ref{sec:decoder}) We qualitatively assess the representation learned by FOCUS, both in simulation and real-world environments. 
\end{itemize}


\textbf{Simulation environments.} We tested our approach in two simulation environments: ManiSkill2 \cite{Gu2023Maniskill2} and robosuite \cite{Zhu2020Robosuite}. 
 For the robosuite environment, we considered a single (red) cube and a two-cube (red and green) setup. For Maniskill2, we opted for two single YCB objects setups (banana and master chef can), and a faucet setup. Segmentation masks for training \OurMethod{}' decoder are provided by the simulator. Find the environments illustrated in Appendix.


\subsection{Supervised dense reward tasks}
\label{sec:dense_tasks}
In Figure \ref{fig:task_results}, we compare the performance in terms of task episodic rewards that are obtained by \OurMethod{} compared to DreamerV2 on supervised dense-reward manipulation tasks. 

As both agents are world model-based agents that learn actor and critic in imagination (and we do not use the object-centric exploration for FOCUS in these experiments), the only difference between the two methods is how the world model is learned, so this study compares the quality of the world model to learn control policies. 

We observe that \OurMethod{} converges faster in the Lift Cube environment and converges to higher rewards in the Stack Cube ($\sim$+10\%) and Turn Faucet ($\sim$+35\%) tasks. This provides empirical evidence that the object-centric model facilitates the learning of dense-reward task behaviors.

\subsection{Exploration and sparse reward tasks}
\label{sec:expl_tasks}

For each environment, we consider a sparse-reward task setting. The tasks analyzed are either ``lift" (L) or ``push" (P) tasks, for cube and YCB objects environments, and ``turn" (T) for the faucet.  The two-stage experimental setup is inspired by the unsupervised RL benchmark \cite{Laskin2021URLB}.

During the \textit{exploration stage}, all exploration approaches (excl. Dreamer), can find the rewards but they do not maximize the environment rewards yet, as they only focus on exploration, using their own exploration rewards. The exploration stage lasts for 2M environment steps. 

During the \textit{adaptation stage}, the agents, pre-trained on the data they collected during the exploration stage, are allowed additional environment steps, aiming to solve the sparse reward tasks, maximizing the reward function. This stage evaluates whether the exploration data was useful to solve downstream tasks.

\textbf{Exploration stage.}
To evaluate the performance of exploration approaches for robotics manipulation, we look into exploration metrics that are related to the interaction with the object and the ability to find potentially rewarding areas in the environment. The metrics we adopt are:
\begin{itemize}
    \setlength\itemsep{0pt}
    \item \textit{Contact (\%):} average percentage of contact interactions between the gripper and the objects over an episode.
    \item \textit{Positional and angular displacement:} cumulative position (m) and angular (rad) displacement of all the objects over an entire episode.
    \item \textit{Up, far, close, left, right placement:} number of times the object is moved in the relative area of the working space. Details about how the areas are defined are in Appendix.
\end{itemize}

\begin{table}[t]
\centering
\resizebox{.85\columnwidth}{!}{%
\begin{tabular}{cccccccccc}
\multicolumn{2}{c}{\multirow{2}{*}{\textbf{}}} &
  \multirow{2}{*}{\textbf{Contact}} &
  \multirow{2}{*}{\textbf{Pos Disp}} &
  \multirow{2}{*}{\textbf{Ang Disp}} &
  \multicolumn{5}{c}{\textbf{Placement}} \\
\multicolumn{2}{c}{}                                   &               &      &               & Up   & ↑     & ↓             & ←     & →             \\ \hline
\multirow{4}{*}{Red cube} &
  \multicolumn{1}{c|}{FOCUS} &
  \textbf{0.04} &
  \textbf{0.15} &
  \textbf{1.86} &
  \textbf{3.5} &
  \textbf{37.45} &
  \textbf{0.06} &
  \textbf{0.15} &
  \textbf{0.01} \\
                        & \multicolumn{1}{c|}{Dreamer} & 0.0           & 0.0  & 0.0           & 0.0  & 0.0   & 0.0           & 0.0   & 0.0           \\
                        & \multicolumn{1}{c|}{P2E}     & 0.0           & 0.0  & 0.01          & 0.0  & 0.15  & 0.0           & 0.0   & 0.0           \\
                        & \multicolumn{1}{c|}{APT}     & 0.0           & 0.01 & 0.15          & 0.01 & 0.11  & 0.0           & 0.0   & 0.0           \\ \hline
\multirow{4}{*}{RG cubes} &
  \multicolumn{1}{c|}{FOCUS} &
  \textbf{0.1} &
  \textbf{0.4} &
  \textbf{4.46} &
  0.0 &
  0.0 &
  0.53 &
  0.0 &
  0.0 \\
                        & \multicolumn{1}{c|}{Dreamer} & 0.0           & 0.01 & 0.01          & 0.0  & 0.0   & 0.16          & 0.0   & 0.0           \\
                        & \multicolumn{1}{c|}{P2E}     & 0.0           & 0.01 & 0.01          & 0.0  & 0.0   & 0.3           & 0.0   & 0.0           \\
                        & \multicolumn{1}{c|}{APT}     & 0.01          & 0.03 & 0.25          & 0.06 & 0.0   & \textbf{5.05} & 0.0   & \textbf{0.03} \\ \hline
\multirow{4}{*}{Faucet*} & \multicolumn{1}{c|}{FOCUS}   & \textbf{0.12} & -    & 1.22          & -    & -     & -             & -     & -             \\
                        & \multicolumn{1}{c|}{Dreamer} & 0.0           & -    & 0.01          & -    & -     & -             & -     & -             \\
                        & \multicolumn{1}{c|}{P2E}     & 0.1           & -    & \textbf{1.23} & -    & -     & -             & -     & -             \\
                        & \multicolumn{1}{c|}{APT}     & 0.08          & -    & 0.98          & -    & -     & -             & -     & -             \\ \hline
\multirow{4}{*}{Banana} &
  \multicolumn{1}{c|}{FOCUS} &
  \textbf{0.08} &
  \textbf{0.38} &
  \textbf{3.5} &
  \textbf{1.31} &
  \textbf{25.28} &
  0.72 &
  \textbf{6.08} &
  \textbf{7.12} \\
                        & \multicolumn{1}{c|}{Dreamer} & 0.0           & 0.0  & 0.06          & 0.04 & 0.22  & 0.02          & 0.16  & 0.29          \\
                        & \multicolumn{1}{c|}{P2E}     & 0.02          & 0.11 & 0.97          & 0.64 & 8.48  & 0.22          & 4.78  & 4.14          \\
                        & \multicolumn{1}{c|}{APT}     & 0.01          & 0.04 & 0.4           & 0.24 & 2.0   & \textbf{0.98} & 1.3   & 2.1           \\ \hline
\multirow{4}{*}{Master Chef Can} &
  \multicolumn{1}{c|}{FOCUS} &
  \textbf{0.05} &
  \textbf{0.49} &
  \textbf{5.78} &
  \textbf{0.17} &
  \textbf{145.66} &
  \textbf{28.57} &
  \textbf{61.51} &
  \textbf{62.8} \\
                        & \multicolumn{1}{c|}{Dreamer} & 0.02          & 0.26 & 3.04          & 0.08 & 85.68 & 24.02         & 24.34 & 46.35         \\
                        & \multicolumn{1}{c|}{P2E}     & 0.03          & 0.29 & 3.26          & 0.12 & 69.95 & 30.63         & 29.72 & 36.0          \\
                        & \multicolumn{1}{c|}{APT}     & 0.01          & 0.14 & 1.56          & 0.06 & 23.64 & 23.74         & 15.88 & 13.52         \\ \hline
\end{tabular}%
}
\vspace{.8em}
\caption{\textbf{Exploration stage results.} Exploration metrics across different environments. *Position displacement and placements are absent for the Faucet as the object is not movable.
}
\label{tab:expl_results}
\end{table}

 In Table \ref{tab:expl_results}, we compare \OurMethod{} against two exploration strategies: Plan2Explore (P2E) \cite{Sekar2020P2E} and a combination of the Dreamer algorithm and Active Pre-training (APT) \cite{Liu2021APT, Rajeswar2023MasteringURLB}. We also test a DreamerV2 agent, which aims to maximize sparse rewards in the environment \footnote{DreamerV2 is still able to explore, with limited capacity, thanks to the actor entropy term in the actor's objective \cite{Hafner2021DreamerV2}.}.

We find that \OurMethod{} shines across all environments, showing strong object interaction performance, and outperforming all other approaches in most metrics. Dreamer manages to find sparse rewards, and thus explore object interactions consistently, only in the master chef can `push' environment, but struggles elsewhere. APT and P2E tend to explore object interactions more consistently than Dreamer, but generally move objects less than \OurMethod{} and tend to find potentially rewarding areas, i.e. achieving placements, more rarely.  Detailed results, showing exploration metrics over timesteps, are presented in Appendix and visual results of FOCUS exploration policies are provided in the \href{https://doubleblindaccount.github.io/}{project website}.

\textbf{Adaptation stage.} After exploring the environment for 2M environment steps, we adapt the exploration approaches, allowing them an additional number of environment interactions. During this stage, if the exploration stage was fruitful, the agent may already know where the rewards can be found in the environment, despite not having actively attempted to maximize them. 

In order to speed up the adaptation stage, task-driven actor and critic networks are pre-trained in imagination, during the exploration stage, using the reward predictor trained on the exploration stage rewards. These actor and critic networks are then fine-tuned during the adaptation stage. This setup is similar to the few-shot adaptation experiments presented in \cite{Sekar2020P2E}.

\begin{figure}[t]
    \centering
    \includegraphics[width=.9\textwidth]{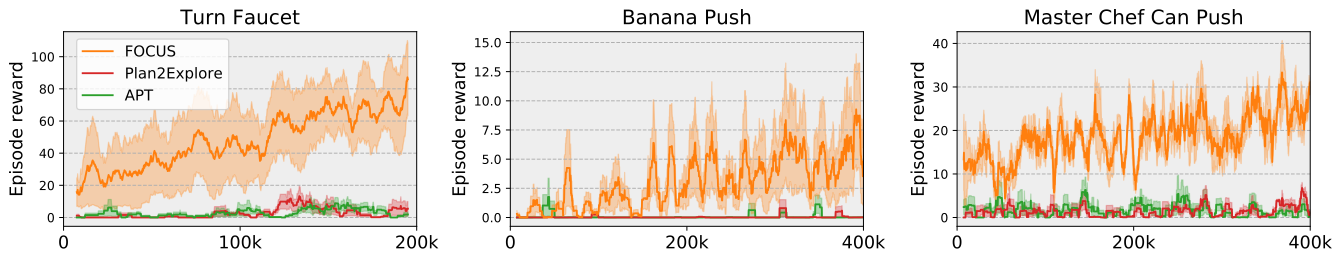}
    \caption{\textbf{Adaptation stage.} Comparing adaptation performance of \OurMethod{} and other exploration strategies in sparse-reward manipulation tasks. (3 seeds)}
    \label{fig:adaptation}
\end{figure}

To provide an idea of how frequently the agents were able to find sparse rewards during the exploration stage, we provide aggregate metrics in Table \ref{tab:task_results}. The adaptation curves, showing episode rewards over time, are presented in Figure \ref{fig:adaptation}. Results clearly show that \OurMethod{} is the only method that makes significant progress across all tasks. As the figures in the Table show, this is mainly due to the fact that \OurMethod{} was also the approach that more frequently found rewards during the exploration stage, making adaptation to downstream tasks easier.



\begin{table}[t]
\centering
\resizebox{\columnwidth}{!}{%
\begin{tabular}{c|ccccccc}
 & Red Cube Push & Red Green Cube Lift & Faucet Turn & \multicolumn{1}{l}{Banana Push} & \multicolumn{1}{l}{Banana Lift} & \multicolumn{1}{l}{MC Can Push} & MC Can Lift \\ \hline
FOCUS   & \textbf{0.67} & 0.0          & \textbf{101.67} & \textbf{38.67} & \textbf{27.0} & 507.67          & \textbf{2.67} \\
Dreamer & 0.0           & 0.0          & 0.0             & 2.67           & 0.67          & \textbf{628.67} & 0.33          \\
P2E     & 0.0           & 0.0          & 35.33           & 19.67          & 14.67         & 207.67          & 2.33          \\
APT     & 0.0           & \textbf{2.0} & 34.33           & 6.67           & 5.33          & 68.33           & 1.67 \\         
\end{tabular}%
} 
\vspace{.8em}
\caption{\textbf{Sparse rewards through exploration.} Summarizing how many times the agents were able to find the sparse rewards during the exploration stage. (3 seeds)}
\label{tab:task_results}
\end{table}

\begin{figure}[t]
\centering
\begin{subfigure}[b]{0.11\textwidth}
    \centering
    \includegraphics[width=\textwidth]{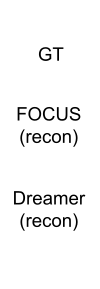}
\end{subfigure}
\hfill
\begin{subfigure}[b]{0.21\textwidth}
    \centering
    \includegraphics[width=\textwidth]{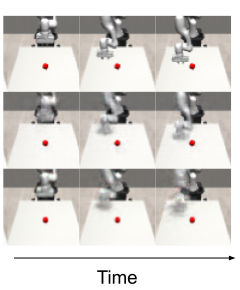}
    \caption{Red cube}
\end{subfigure}
\hfill
\begin{subfigure}[b]{0.21\textwidth}
    \centering
    \includegraphics[width=\textwidth]{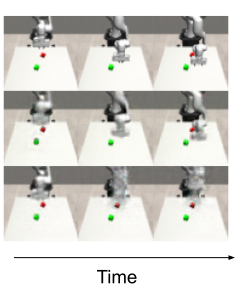}
    \caption{RG cubes}
\end{subfigure}
\hfill
\begin{subfigure}[b]{0.21\textwidth}
    \centering
    \includegraphics[width=\textwidth]{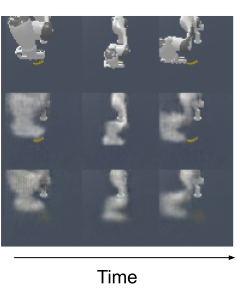}
    \caption{Banana}
\end{subfigure}
\hfill
\begin{subfigure}[b]{0.21\textwidth}
    \centering
    \includegraphics[width=\textwidth]{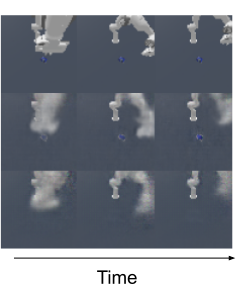}
    \caption{Master chef can}
\end{subfigure}
    \caption{\textbf{Simulation reconstructions.} Comparison of Dreamer and FOCUS reconstructions on random action trajectories. }
    \label{fig:reconstructions}
\end{figure}
\subsection{Model Analysis}
\label{sec:decoder}

\textbf{Simulation.} We present results in the simulated environments, sampling random action trajectories, in Figure \ref{fig:reconstructions}. 
In order to qualitatively assess the capacity of \OurMethod{} to better capture information about the objects, we compare reconstructions from \OurMethod{} to the non-object-centric world model of DreamerV2. We observe that, while Dreamer can reconstruct the robot and the background, it struggles to accurately reconstruct objects. The banana is reduced to a yellow spot, while the can is almost invisible against the background. \OurMethod{}, instead, reconstructs the objects more accurately, by capturing the object information in the object latents.

\begin{figure}[b]
    \centering
    \includegraphics[width=.9\textwidth]{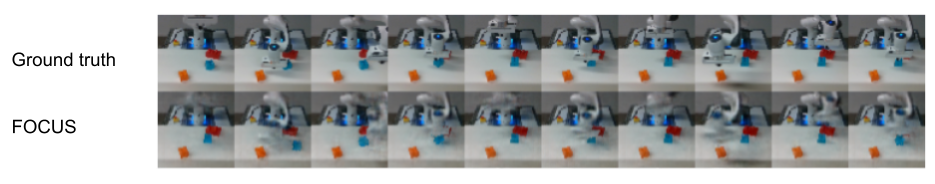}
    \caption{\textbf{Real world reconstructions.} Comparing ground truth and FOCUS reconstructions on real-world sequences of data.}
    \label{fig:rw_reconstructions}
\end{figure}

\textbf{Real-world.} In order to show that \OurMethod{} can also be applied to real-world setups, we present results in a real-world setting, where we use a Franke Emika arm in front of a white table with three colored bricks (blue, orange, red). The main issue with applying \OurMethod{} in the real world comes from the absence of segmentation masks. However, thanks to recent progress in the study of large-scale vision models, the issue can be easily circumvented using a pre-trained segmentation model. For these experiments, we adopted the Segment Anything Model (SAM) \cite{Kirillov2023SAM}, which we found particularly proficient at providing segmentation masks and tracking objects, after providing a single labeled sample to recognize the bricks in the scene. In Figure \ref{fig:rw_reconstructions}, we show that our method effectively captures object information, though sometimes sacrificing information about the gripper. We saw this happening for the simulated environments as well (Figure \ref{fig:reconstructions}), nonetheless, this did not affect performance negatively. We believe the reason for that is that the agent can exploit the proprioceptive information for control. However, if the agent does not capture information about objects correctly, this can significantly harm the task's performance.   

\section{Conclusion} 
\label{sec:conclusion}
We presented \OurMethod{}, an object-centric model-based agent for robotics manipulation.
Our results show that our approach can learn manipulation tasks more efficiently, thanks to the object-centric world model, and explore robot-object interactions more easily, thanks to a novel object-driven exploration reward. As we showed, this can facilitate solving even challenging sparse rewards tasks, where finding rewards can be hard and require demonstrations or task-specific strategies. 
In the future, we aim to develop our object-centric representation further, aiming to remove dependence upon the segmentation information, used in the object decoder's loss, which is not easily available to a real robot. In order to overcome such limitations, we aim to investigate unsupervised strategies for scene decomposition \cite{Wu2023SlotFormer, Locatello2020SlotAttention}.



\section{Limitations}

\OurMethod{} requires segmentation masks over the object of attention in order to function. Segmentation masks can be challenging to obtain in real-world scenarios. To overcome this limitation, we developed a segmentation pipeline based on SAM \cite{kirillov2023segment, cheng2023segment}. 

In our current implementation, every object of interest is uniquely identified by a one-hot vector $c^{obj}$. This requires prior knowledge about the number and the identity of objects in the scene. In the future, this should be addressed by relying on more sophisticated sets of object features, which would allow us to identify objects, before feeding the information to the object decoder.

Reconstruction-based models may struggle with changing dynamics in the environment or complex backgrounds \cite{Stone2021distractingCS}, making the adoption of world models challenging in real-world scenarios, where the environment is often uncontrolled. To address this limitation in the future, we aim to remove the background from the reconstruction scope, focusing solely on the objects in the scene and on the proprioceptive information, or using some alternative learning paradigms, e.g. contrastive learning \cite{Chen2020SimCLR}, to reduce the amount of information stored in the representation about the background.


\clearpage
\acknowledgments{This research received funding from the Flemish Government (AI Research Program). Pietro Mazzaglia is funded by a Ph.D. grant of the Flanders Research Foundation (FWO).}


\bibliography{main}  

\newpage
\clearpage

\appendix

\section{Model details}

\textbf{Dreamer.} The architecture adopted for Dreamer is relevant to the one documented by \cite{Hafner2021DreamerV2}. Model states have both a deterministic and a stochastic component: the deterministic component is the 200-dimensional output of a GRU (\cite{Cho2014GRU}, with a 200-dimensional hidden layer; the stochastic component consists of 32 categorical distributions with 32 classes each. States-based inputs such as the proprioception, the encoder, and the decoder are 4-layer MLP with a dimensionality of 400. For pixels-based inputs, the encoder and decoder follow the architecture of Dreamer \cite{Hafner2021DreamerV2}, taking 64 × 64 RGBD images as inputs. Both encoder and decoder networks have a depth factor of 48. To ensure stable training during the initial phases, we adopt a technique from \cite{Hafner2023DreamerV3} where the weights of the output layer in the critic network are initialized to zero. This approach contributes to the stabilization of the training process especially in the early stages of training.

Networks are updated by sampling batches of 32 sequences of 32 timesteps, using Adam with learning rate $3\mathrm{e}^{-4}$ for the updates, and clipping gradients norm to 100.

\textbf{\OurMethod{}.} The architecture proposed is based on the implementation of Dreamer described above. The encoding network and the state-based decoding unit have the same structure mentioned in Dreamer. We introduced an object latent extractor unit consisting of a 3-layer MLP with a dimensionality of 512. The object-decoder network resembles the structure of the Dreamer's decoder, the depth factor for the CNN is set to 72. 64x64 RGBD images along with a "segmentation weights" image are generated per each object.

\textbf{Objective.} In the objective of \OurMethod{} we describe object-specific masks $m^i_t$. These masks are binary images, obtained from the entire scene segmentation mask. In practice, $m_t = \sum_i{m^i_t}\cdot \arg\max{c^i}$, which means the overall segmentation mask is obtained by summing the object-specific masks, multiplied by their object index, which can be extracted from $c^i$ through an $\arg\max$, being $c^i$ a one-hot vector.

\section{Environment details}

\begin{figure*}[t]
\begin{subfigure}[b]{0.195\textwidth}
    \centering
    \includegraphics[width=\textwidth]{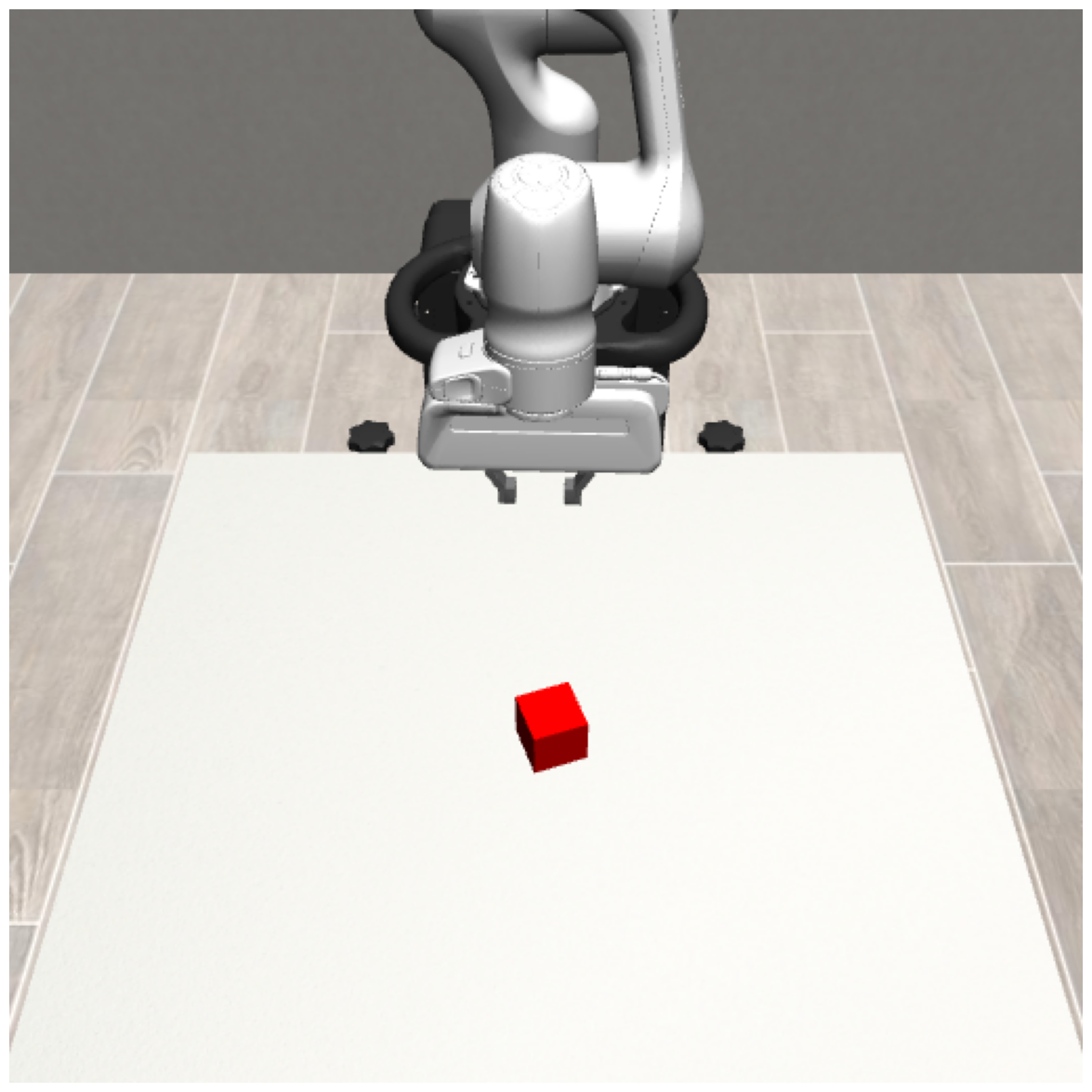}
    \caption{Red cube}
\end{subfigure}
\hfill
\begin{subfigure}[b]{0.195\textwidth}
    \centering
    \includegraphics[width=\textwidth]{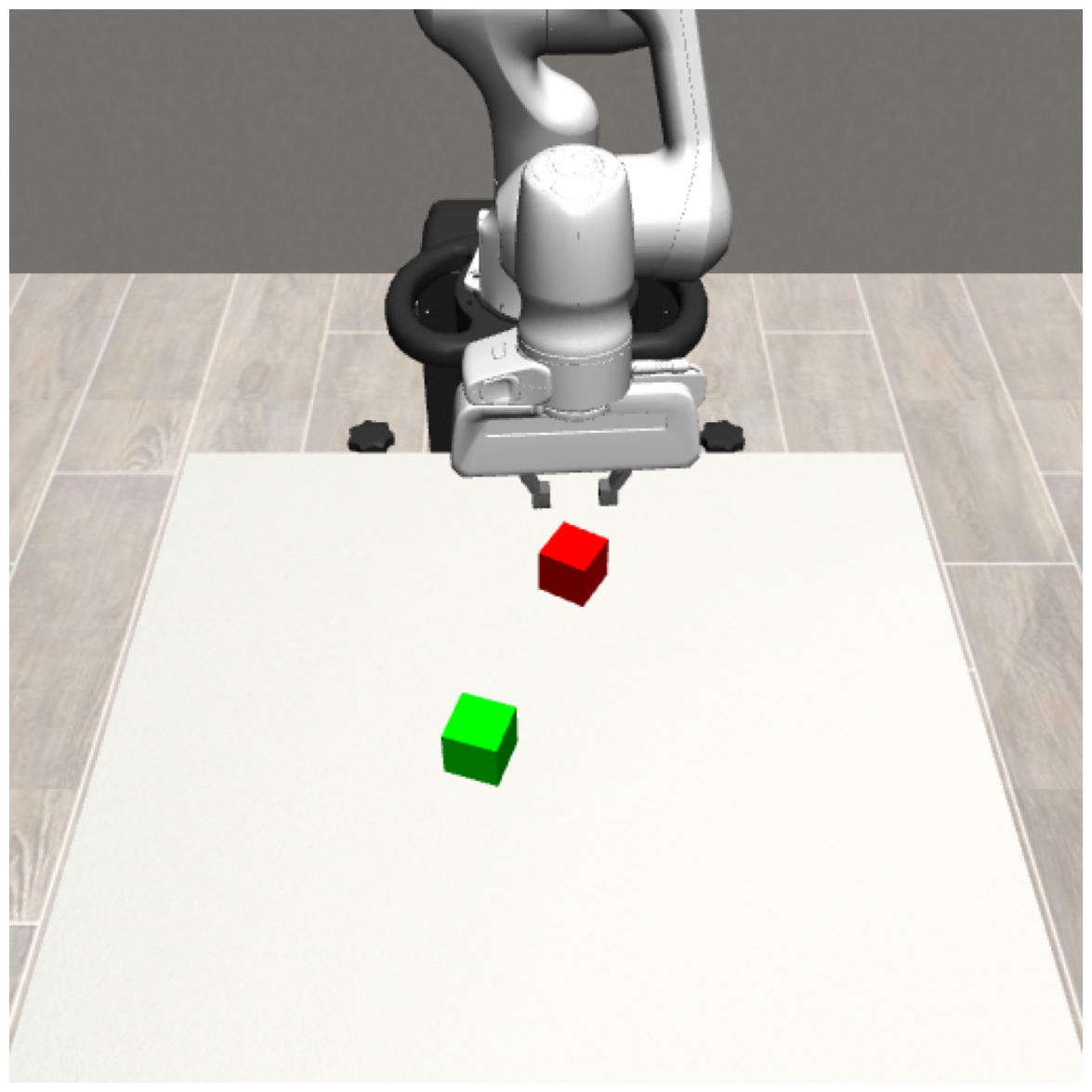}
    \caption{RG cubes}
\end{subfigure}
\hfill
\begin{subfigure}[b]{0.195\textwidth}
    \centering
    \includegraphics[width=\textwidth]{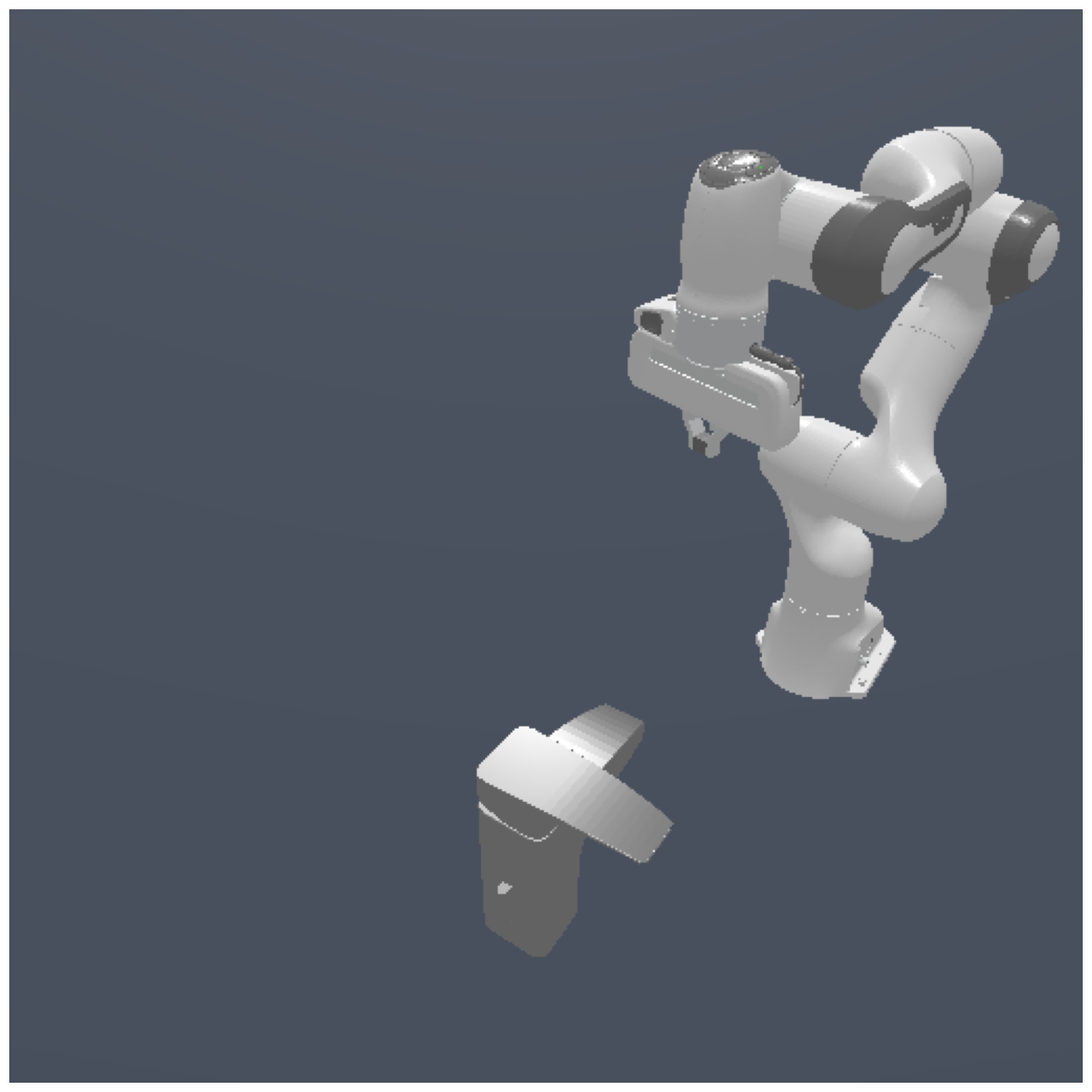}
    \caption{Faucet}
\end{subfigure}
\hfill
\begin{subfigure}[b]{0.195\textwidth}
    \centering
    \includegraphics[width=\textwidth]{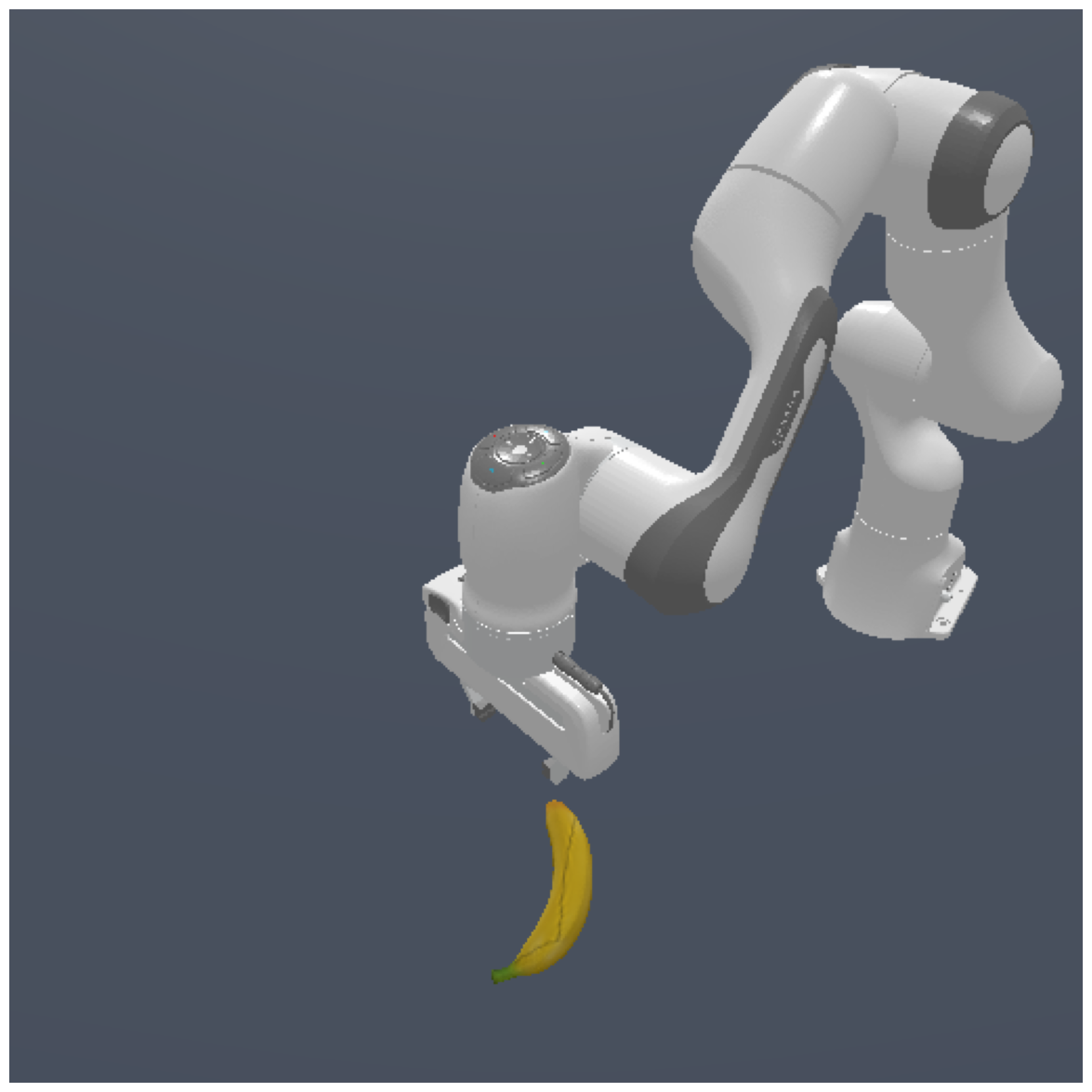}
    \caption{Banana}
\end{subfigure}
\hfill
\begin{subfigure}[b]{0.195\textwidth}
    \centering
    \includegraphics[width=\textwidth]{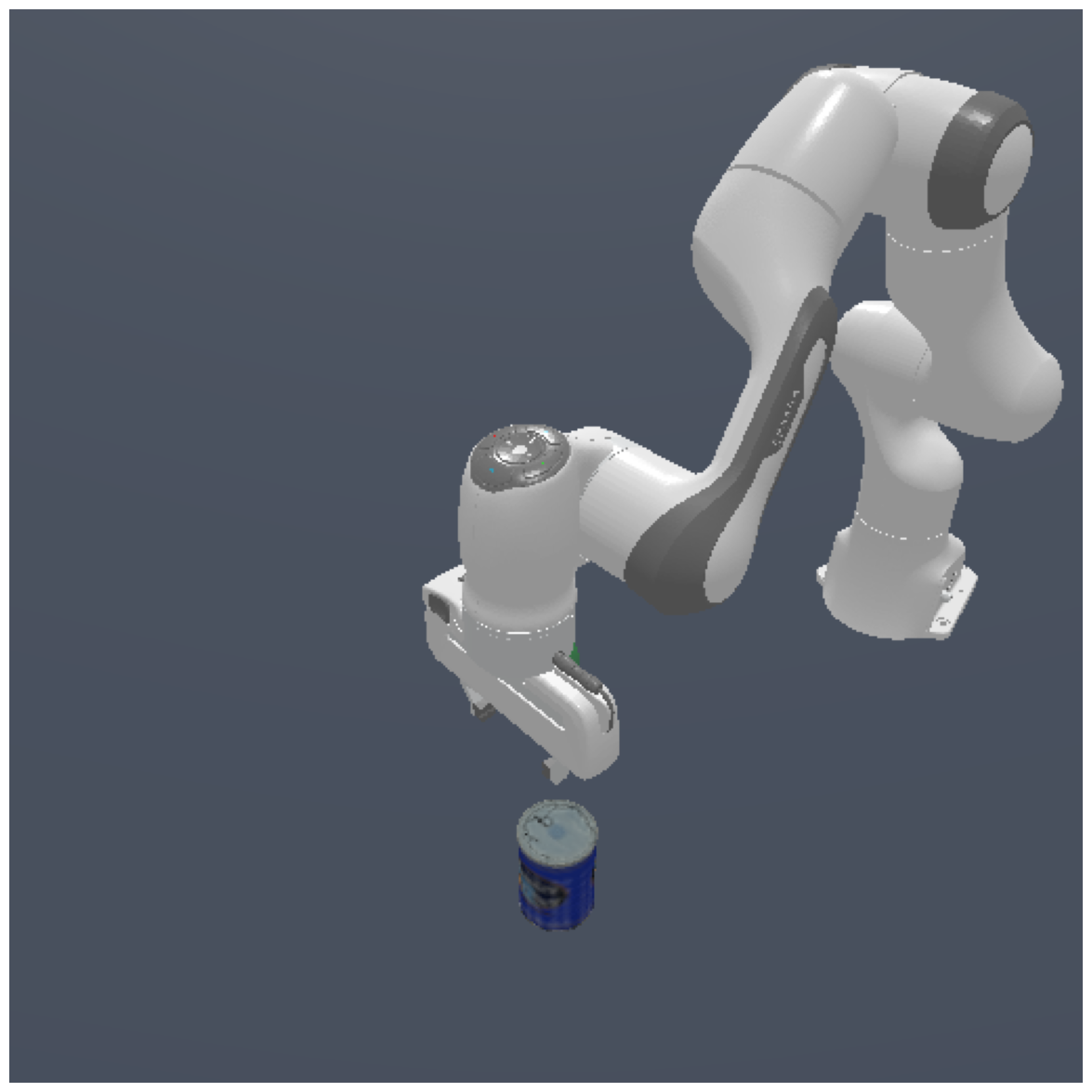}
    \caption{Master Chef Can}
\end{subfigure}
\caption{\textbf{Environments.} Visualization of the environments we applied \OurMethod{} on. \textit{(a-b)} from robosuite. \textit{(c-e)} from ManiSkill2.}
\label{fig:environments}
\end{figure*}

In both simulation settings, ManiSkill and robosuite, the robotic agent is a Franka arm with 7-DoF, controlled in cartesian end-effector space with fixed gripper orientation. Gripper state (open/close) adds a degree of freedom.
For all the mentioned environments objects are spawned at a fixed pose. Visual observations are rendered at 64x64 resolution for all the environments. Considered tasks are shown in figure \ref{fig:environments}. 

\subsection{Metrics and task rewards}
To evaluate the placement metric, we divide the workspace in front of the robot into 5 areas, with respect to the 5 cited directions: right, left, far, close, and up. For the Robosuite environment, with respect to the origin of the environment (coinciding with the center of the table) we consider a successful placement if the object is placed at a minimum of 0.25m up to a maximum of 0.4m along the directions considered. For the "up" placement the minimum threshold is reduced to 0.05m.

For the Maniskill2 environment, in the same logic, we considered a minimum threshold of 0.4m up to a maximum of 0.5m. For the "up" placement a minimum threshold is reduced to 0.1m.

The definition of the task rewards for the exploration tasks is linked to the placement definition given. Any reward is provided till the object has been placed in the desired area.  
For the lift tasks, before assigning any reward we check if the object is grasped.
The definition of the task rewards for dense tasks is the one defined by the default implementation.

\clearpage
\newpage

\section{Extended results}

\begin{figure}[hbt!]
\centering
\begin{subfigure}[b]{\textwidth}
    \centering
    \includegraphics[width=\textwidth]{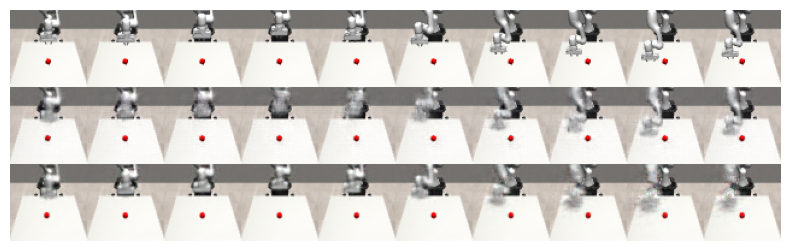}
    \caption{Red cube}
\end{subfigure}
\hfill
\hfill
\begin{subfigure}[b]{\textwidth}
    \centering
    \includegraphics[width=\textwidth]{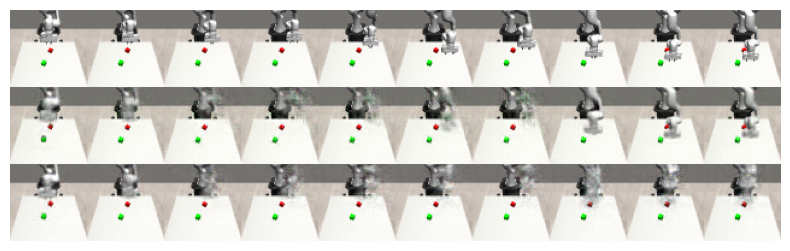}
    \caption{RG cubes}
\end{subfigure}
\hfill
\hfill
\begin{subfigure}[b]{\textwidth}
    \centering
    \includegraphics[width=\textwidth]{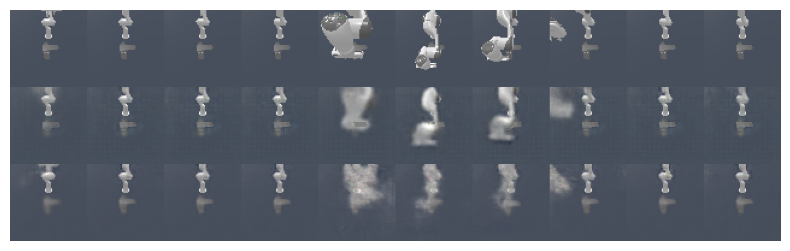}
    \caption{Faucet}
\end{subfigure}
\end{figure}

\clearpage

\begin{figure*}[t!]
\ContinuedFloat
\hfill
\begin{subfigure}[b]{\textwidth}
    \centering
    \includegraphics[width=\textwidth]{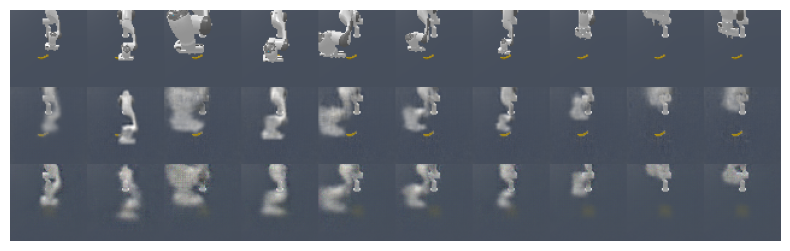}
    \caption{Banana}
\end{subfigure}
\hfill
\centering
\begin{subfigure}[b]{\textwidth}
    \includegraphics[width=\textwidth]{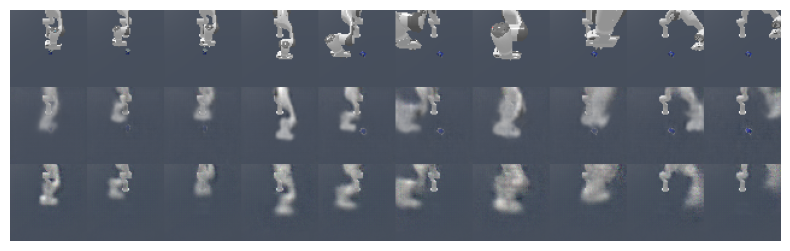}
    \caption{Master Chef Can}
\end{subfigure}
\caption{\textbf{Model reconstructions over random actions for simulated environments.} First row is ground truth. Second row is FOCUS reconstruction, overlap of instance masks. Last row shows Dreamer reconstruction.}
\vspace{100in}
\label{fig:extended_visualizations}
\end{figure*}

\clearpage

\begin{figure*}
    \vspace{-4em}
    \centering
    \includegraphics[width=\textwidth]{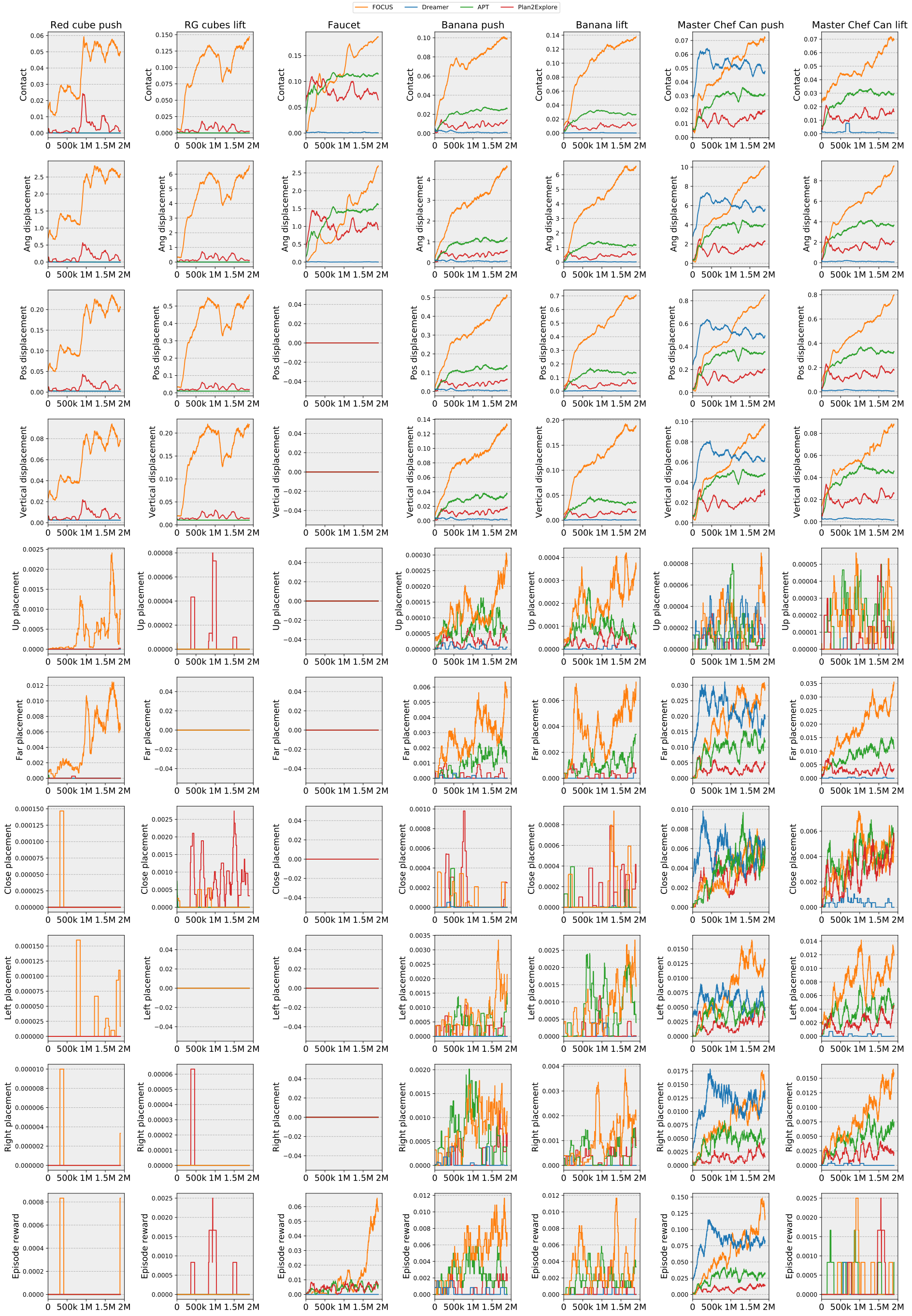}
    \caption{Displaying exploration metrics over time for the exploration experiments.}
    \label{fig:extended_exploration}
\end{figure*}

\end{document}